\documentclass[letterpaper, 10 pt, conference]{ieeeconf}  %

\IEEEoverridecommandlockouts                              %
\usepackage[l2tabu,orthodox]{nag}

\usepackage[
backend=biber, 
bibencoding=utf8,
style=ieee, 
sorting=none, 
natbib=true, 
doi=false, 
isbn=false, 
url=false, 
eprint=false, 
maxcitenames=1, 
mincitenames=1,
]{biblatex}

\usepackage[draft,pdftex,colorlinks]{hyperref}
\usepackage[printonlyused]{acronym}

\usepackage{siunitx}
\usepackage[all]{nowidow}

\usepackage[dvipsnames]{xcolor}

\usepackage{lipsum}

\usepackage[pdftex]{graphicx}

\usepackage{epstopdf}

\usepackage{import}

\graphicspath{{./latexGoodPractices/}}

\usepackage{booktabs}

\usepackage{tabu}

\usepackage{amssymb,amsfonts,amsmath,amscd}

\usepackage{bm}

\newcommand{\bbm}{\begin{bmatrix}}
\newcommand{\ebm}{\end{bmatrix}}

\usepackage[utf8]{inputenc}
\usepackage{commath}
\usepackage{mathtools}
\usepackage{siunitx}
\usepackage{algpseudocode}
\usepackage[ruled,vlined]{algorithm2e}
\usepackage{framed}
\usepackage{color,soul}

\graphicspath{{media/}}

\newcommand{\ie}{i.e., }
\newcommand{\eg}{e.g., }

\renewcommand{\P}[2][]{\mathbb{P}_{#1}\left(#2\right)}
\newcommand{\E}[1]{\mathbb{E}\left[#1\right]}
\newcommand{\R}{\mathbb{R}}
\newcommand{\expp}[1]{\exp\left(#1\right)}
\newcommand{\area}{\Delta a}

\acrodef{TTC}{Time To Collision}

\usepackage{soul}

\title{\LARGE \bf
	Dynamic Lambda-Field: A Counterpart of the Bayesian Occupancy Grid for Risk Assessment in Dynamic Environments
}

\author{Johann Laconte$^1$, Elie Randriamiarintsoa$^1$, Abderrahim Kasmi$^{1,2}$, François Pomerleau$^{3}$, \\
	Roland Chapuis$^1$, Christophe Debain$^4$, Romuald Aufrère$^1$
\thanks{$^{1}$ Universit\'e Clermont Auvergne, CNRS, Clermont Auvergne INP, Institut Pascal, F-63000 Clermont-Ferrand, France; johann.laconte@uca.fr}%
\thanks{$^{2}$ Sherpa Engineering, R\&D Department, 333 Avenue Georges Clemenceau, 92000 Nanterre, France; a.kasmi@sherpa-eng.com}
\thanks{$^{3}$ Northern Robotics Laboratory, Universit\'e Laval, Canada; francois.pomerleau@ift.ulaval.ca}%
\thanks{$^{4}$ Université Clermont Auvergne, INRAE, UR TSCF, F-63178 Aubi\`ere, France; christophe.debain@inrae.fr}%
}

\usepackage[switch]{lineno}

\begin{document}

\maketitle
\thispagestyle{empty}
\pagestyle{empty}

\begin{abstract}
	In the context of autonomous vehicles, one of the most crucial tasks is to estimate the risk of the undertaken action.
	While navigating in complex urban environments, the Bayesian occupancy grid is one of the most popular types of maps, where the information of occupancy is stored as the probability of collision.
	Although widely used, this kind of representation is not well suited for risk assessment:
	because of its discrete nature, the probability of collision becomes dependent on the tessellation size. %
	Therefore, risk assessments on Bayesian occupancy grids cannot yield risks with meaningful physical units.
	In this article, we propose an alternative framework called Dynamic Lambda-Field that is able to assess generic physical risks in dynamic environments without being dependent on the tessellation size. %
	Using our framework, we are able to plan safe trajectories where the risk function can be adjusted depending on the scenario.
	We validate our approach with quantitative experiments, showing the convergence speed of the grid and that the framework is suitable for real-world scenarios.

\end{abstract}

\section{Introduction}

In the context of robotics in urban environments, autonomous vehicles are beginning to coexist alongside human drivers.
One of the main factors to take into account for each robot's decision is undoubtedly the risk of the undertaken action.
The risk is tied to the probability of collision, conventionally defined as the probability of encounter between the robot and static or dynamic obstacles.

To assess this risk, the robot creates a map of the environment and identifies any potential threats or obstacles it must avoid.
To create these maps, the robot can either use a semantic approach, namely identifying each obstacle as a unique entity, or use a metric approach by storing the occupancy information of each position in the environment.
Although the first approach seems by far the most fitted for risk assessment scenarios, creating semantic maps is not an easy task.
Indeed, identifying obstacles from raw measurements is complicated and can leave obstacles undetected if, for example, the obstacle has specific features that were not in the training set (\eg people wearing a skirt or with strollers \cite{Kidono2011}).
As it is critical to detect any possible threat, the metric map is then a reliable option since it does not require in-depth processing of measurements, thus providing a more conservative approach to risk assessment.

\begin{figure}[t]
	\centering
	\includegraphics[width=\linewidth]{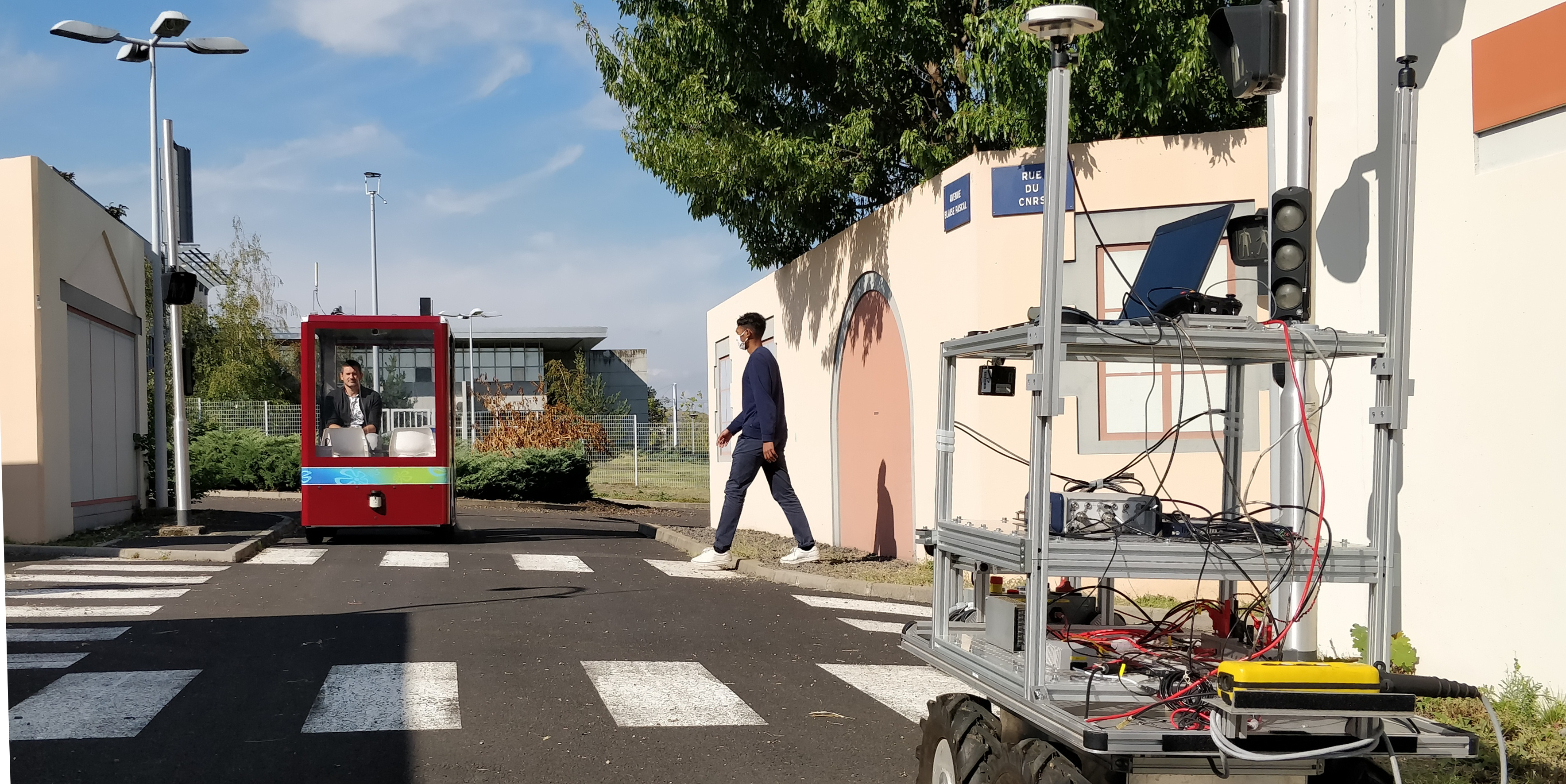}
	\caption{Example of an urban scenario that the robot can encounter. A pedestrian is crossing the road, while a vehicle is approaching the robot in the other lane.
			 Using the Dynamic Lambda-Fields, the robot is able to estimate the risk of its actions, taking into account a generic risk that can be adapted depending on the scenario.
		}
	\label{fig:intro}
	\vskip-3em
\end{figure}
The most commonly used metric map is the Bayesian occupancy grid: %
it tessellates the environment into cells where each cell stores the probability of occupancy (\ie the probability that the cell will cause a collision if crossed). 
However, as highlighted by \citet{Heiden2017}, this type of map is not suitable for risk assessment.
Indeed, the probability of collision depends on the size of the cells: for instance, for a map having cells twice as small as another map of the same environment, the robot would have to cross twice as many cells to reach its goal, increasing the probability of collision.
Following this problem, any risk assessment strategy on Bayesian occupancy grids has to take into account the cell size in the equations.
As such, the Bayesian occupancy grid cannot naturally yield risks but only probabilities of collision that are themselves dependent on the tessellation size.
For instance, \autoref{fig:intro} shows a scenario where a robot is crossing an urban-like environment: a car is coming in the other lane while a pedestrian suddenly crosses the road despite the vehicles approaching.
In the case where the dynamics of the robot does not allow it to stop, a collision is inevitable, either with the pedestrian by simply braking, or with the vehicle in the other lane by swerving.
Whereas Bayesian occupancy grid frameworks are not able to differentiate by themselves the risks associated with each decision, a human driver would have a more nuanced approach.
If the risk only takes into account the damages to the robot itself, it is indeed safer to collide with the pedestrian since its mass is much lower than that of a car.
However, if the risk takes into account the probability that both the robot and the collided obstacle survive, changing direction and colliding with the car might become a better solution.

In this article, we present a novel framework, named Dynamic Lambda-Field, specifically conceived to allow risk assessment in metric maps using a 2D lidar sensor.
This concept answers the limitations of the Bayesian occupancy grid by proposing a method where the probability of collision does not depend on the tessellation size, thus allowing for continuous risk assessments over the path of the robot.
We extend our theory presented in {\cite{Laconte2019}} to take into account dynamic environments.
Using the framework presented in this article, the robot is able to plan safe trajectories using generic, user-defined risks that do not depend on the tessellation size while keeping their physical units.

\section{Related Work}
The most widely adopted representation of static environments is the Bayesian occupancy grid, introduced by \citet{Elfes1989}.
\citet{Negre2014} proposed to extend the framework, estimating static and dynamic obstacles in occupancy grids, using particles to represent the dynamic world.
\citet{Rummelhard2015} enhanced the aforementioned framework by conducting a pre-processing step of estimating the state of the space (\ie static, dynamic, empty or unknown).
Our framework draws from this idea, inferring at each step the state of each cell, while providing a means of inferring physical risks from the map. %
In parallel, \citet{Nuss2016a} proposed to model the dynamic environment using random finite sets and Dempster-Shafer theory.
\citet{Rexin2017} built upon the previous method, using a pure Bayesian approach, leading to analogous results \cite{Rummelhard2015}. %
\citet{Guizilini2019} proposed a method to learn and store continuous dynamic occupancy maps. 
Although providing an approach to learn complex pattern and dependencies in the environment, \cite{Guizilini2019} relies on clustering and learning, two aspects where providing safety guaranties cannot be ensured. %
Furthermore, the problem of the tessellation size remains in continuous maps: the question only becomes how to integrate (instead of sum) probabilities of occupancy.
As the information stored into such maps is not suitable for risk assessment, the Bayesian occupancy grid cannot yield meaningful risks.

The notion of risk and how to handle it is not uniquely defined in the literature. %
\citet{Majumdar2017} addressed this question, exploring what criteria are needed to define risk metrics.
In the context of occupancy grids, \citet{Rummelhard2014} proposed to define the risk as the probability to collide with a specific area.
The authors argued that a simpler approach such as the maximum value of collision over the cells tends to work better.
Nevertheless, this approach discards all collisions with the exception of the one with the maximum likelihood, leading to potential hazardous situations (\eg a path colliding with two pedestrians with chances of \SI{90}{\%} and \SI{91}{\%} may be more dangerous than a path colliding with only one pedestrian with a probability of \SI{95}{\%}).
\citet{Guardini2020} inferred risk maps from a Bayesian occupancy grid and defined it as the severity of injury.
However, they do not provide a means of inferring the risk of a path as the aforementioned problem of the Bayesian maps also applies here.
As such, our framework can be seen as a counterpart of their work where it is possible to assess the risk of a path.
\citet{Heiden2017} proposed a method to assess collision probabilities in Bayesian occupancy grids, relying on product integrals.
The main drawback is that there is no natural reason to use the concept of product integrals, as it is here merely a tool to lift the dependence on the tessellation size. %
In addition, their framework does not take into account the size of the robot.
On the contrary, our method takes into account its size while providing a natural means of assessing generic risks.
Finally, \citet{Fan2021} planned trajectories in risk maps (\ie a grid where each cell stores a risk) and defined a risk as a compound of one-step risk metrics.
Each risk is defined as a weighted sum of different metrics (\eg collision, slippage).
However, compounding and summing risks therefore lose their physical meaning, whereas our framework maintains the physical meaning and therefore leads to more informed decisions.
In a different manner, \citet{Laugier2011} clustered the dynamic obstacles of the Bayesian occupancy grid, hence falling back on more standard path planning problems where the obstacles are identified.
However, this approach assumes that every obstacle has been successfully detected.
Because of the possible wrong clustering (\eg two pedestrians being detected as only one cluster), obstacles might have an incorrect velocity estimation hence leading the robot to make wrong decisions.
On the contrary, our method does not require performing clustering to assess risk with a physical meaning, hence provides a more conservative approach.
The risk function, that we define below as the maximum change of kinetic energy arising from a collision, keeps its physical unit and allows more nuanced decision-making than would be possible with only the probability of collision.
Therefore, more adequate decisions can be taken by the robot, where each path possesses a meaningful risk with a physical unit. %

\section{Risk assessment in Dynamic Lambda-Field}
	We present here how to construct Dynamic Lambda-Fields as well as how to assess risks using this map as done in the static case in \cite{Laconte2019}.
In order to motivate our approach, we show that the theory of the Lambda-Field naturally emerges from Bayesian occupancy grids when the cell size tends to zero.
We assume that when the cell size tends to zero, the probability of a collision occurring in this cell is $\lambda\area$, where $\lambda\in\R_{\geq 0}$ is the rate of the event collision and $\area$ is the area of the cell.
The larger $\lambda$ is, the more likely a collision will occur.
Under these assumptions, the probability to safely cross a path (\ie without collision) of $N$ cells $c_i$ of areas $\area$ with a constant rate of $\lambda_i$ is
\begin{equation}
	\prod_{i=0}^{N-1} (1-\lambda_i\area).
\end{equation}
Taking the limit $\area\!\! \rightarrow\!\! 0, N\!\!\rightarrow\!\! \infty$ leads to the computation of the Volterra type I product integral.
For a path crossing a total area $A$ where each cell of area $\area$ has a rate $\lambda(a)$, $a$ being the total area crossed from the beginning, the probability of collision is
\vskip-1em
\begin{equation}
\begin{aligned}
	\P{\mathtt{coll}} &= 1 - \lim_{\area\rightarrow 0}\prod_{i=0}^{A/\area} \left(1-\lambda(i\area)\area\right) \\
					  &= 1 - \expp{-\int_0^A\lambda(a)\dif a}.
\end{aligned}
\end{equation}
It can be shown that taking such a limit leads to the Poisson point distribution.
Therefore, the theory of Lambda-Field relies on this distribution.
Instead of storing the probability of occupancy for each obstacle (\ie cell and potential dynamic obstacles), as done in the standard Bayesian occupancy grids, the Lambda-Field stores the intensity of the cells, namely a value $\lambda_i \in \R_{\geq 0}$ for each obstacle.
The higher the intensity, the higher is the probability that crossing the cell will result in a collision.
Conversely, a value of zero indicates that the cell will never lead to a collision.

In this article, we define the risk as the maximum change of kinetic energy of the robot and the obstacle (static or dynamic) due to a collision. %
The framework allows the risk function to be changed depending on the application.
For example, the risk function could also model the probability of survival for both the vehicle occupants and the collided obstacle if it is human.
By keeping the risk in its physical form, the robot is able to make more informed decisions without the need of tuned user-defined thresholds.
First, we show how to compute the Lambda-Field, namely the intensities $\lambda_i$ of both static and dynamic obstacles, using a lidar sensor. 
Then, we present how to manage the particles used to represent the obstacles in the environment.
Finally, we show how to use this field to assess physical risks given a generic risk function, leading to the generation of safe paths for the robot.

\subsection{Mapping}
	As the environment contains both static and dynamic obstacles, the value of the lambda of a cell $c_i$ (\ie its intensity representing the likelihood of a collision occurring in this cell) is stochastic.
	Indeed, the cell can be occupied by a static obstacle, a dynamic obstacle or be free of obstacles.
	Therefore, we define the probability of collision as the probability of colliding with an obstacle in the expectation of the intensity field, leading to
	\vskip-1em
	\begin{equation}
	\begin{aligned}
		\P{\mathtt{coll}} = 1 - \expp{-\sum_{c_i\in\mathcal{C}} \area \E{\lambda_{c_i}^{t_i}}},
	\end{aligned}
	\label{eq:field}
	\end{equation}
	for a path crossing the cells $\mathcal{C} = \{c_0, \dots, c_{N-1}\}$ at the times $\{t_0, \dots, t_{N-1}\}$, of expected intensities $\left\{\E{\lambda_{c_0}^{t_0}}, \dots, \E{\lambda_{c_{N-1}}^{t_{N-1}}}\right\}$ and of area $\area$.
	One can note that the probability of collision does not depend on the tessellation size as the sum is weighted by the area of the cells, as shown in \cite{Laconte2019}.

	In order to estimate the expectation of the intensity $\E{\lambda_c^t}$ for each cell {$c$}, we use particles that represent the possible obstacles in the environment.
	The static grid is considered as a set of particles that do not move and have the size of a cell, whereas dynamic obstacles are defined as moving particles of different classes.
	We define in this work three different classes of particles: the `cell' class, for the static environment, the `pedestrian' class and the `car' class for the dynamic obstacles.
	Each particle has a position, a speed, a velocity profile (\ie a car can achieve greater speed than a pedestrian, whereas a pedestrian can change direction quicker).
	Each particle is also defined by their size. 
	For instance, a pedestrian particle is represented by a \SI{40}{\cm} diameter circle whereas a vehicle is described by a rectangle of $\SI[parse-numbers=false]{2\times1}{\m}$. 
	Note that the particles' footprints are not dependent on the tessellation size.
	Naturally, particles of the `cell' class have null velocity and acceleration.
	Using these attributes, the framework is able to represent the different obstacles more efficiently, as only one particle can represent an obstacle even if it spans several cells, contrary to {\cite{Rummelhard2015}} where a particle does not have a size.
	Also, using the particle classes representing the obstacle, the classes of the obstacles are inferred at the same time, allowing the framework to take into account the classes of the obstacles while assessing the risk of a path.
For instance, if the risk function takes into account the probability of survival of collided obstacles, it allows the robot to prefer colliding with a car over a pedestrian.

\begin{figure}[htbp]
		\centering
		\includegraphics[width=\linewidth]{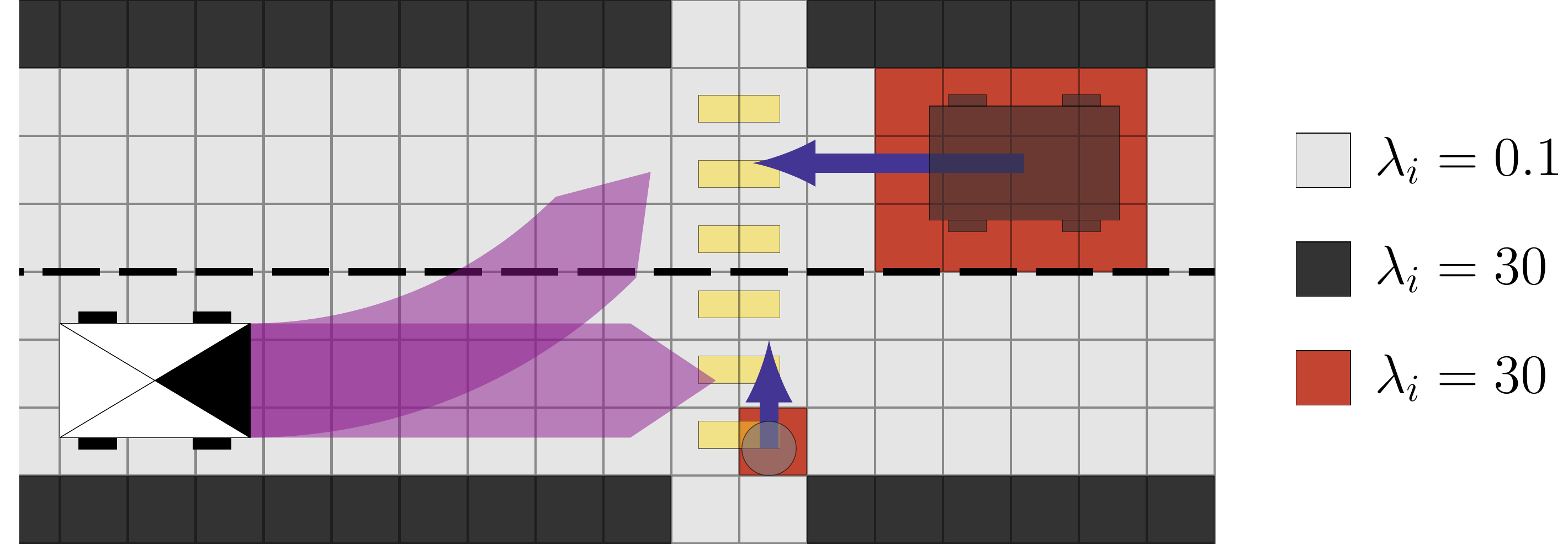}
		\caption{Example of Dynamic Lambda-Field where a pedestrian emerges unexpectedly on the road in front of the robot (black box with its front represented as a filled triangle) while another vehicle is approaching in the opposite lane.
			The environment is represented as a set of cells storing the expected intensity, computed using the intensity of the static (resp. dynamic) particles, depicted in gray (resp. red) scale, occupying these cells.
			Depending on the risk function, the robot can assess which one of the two paths (in purple) is the safest.
				}
		\label{fig:example_lf}
	\end{figure}

	As an example, \autoref{fig:example_lf} gives the Lambda-Field of \autoref{fig:intro} where the robot wants to cross the environment while a pedestrian abruptly appears on the road, therefore having only two choices: either braking and colliding with the pedestrian or swerving and colliding with the car in the other lane, hence saving the pedestrian.
	The environment is populated by static particles (one per cell) of different intensities whereas the possible dynamic obstacles are represented using dynamic particles (in this example, only two, one of type `pedestrian' and the other of type `car').
	The expectation of the cells' intensity that the robot crosses is derived in the following section, taking into account the intensity of the particles that are in the cell at the time of traversal (a static particle and possibly several dynamic ones).
	Once the expected intensity of the cells is computed, the probability of collision of the robot's path is given by \autoref{eq:field}, integrating the expected intensity of the cells crossed by the robot.
	Then, the expectation of a risk function over the path can be computed, giving more insight on which path is the safest.

\subsection{Computation of the expectation of a cell}
	In order to measure particles and infer the risk of a path, the first step is to compute the expectation of the intensities of the cells.
	When measuring the environment at a given time, we assume that a cell cannot contain more than one obstacle (\eg a cell can contain a wall or a car but not both).
	This consideration allows us to only measure the obstacle that is in the cell, as for instance we do not want to measure the static part of a cell as occupied if a moving car is in the same cell at this time.
	Furthermore, we also model the probability that a particle still exists: it is indeed unlikely that a particle still represents an obstacle after having run into a wall.
	If this case occurs, the existence probability of the particle will drop.
		
	Under these considerations, the expected lambda of a cell $c$ is given by
	\begin{equation}
		\E{\lambda_c^t} = \sum_{p_i\in c}\lambda_i\P{p_i\in c}\P{ e_i \,\middle|\, p_i\in c},
		\label{eq:e_lambda}
	\end{equation}
	where $\lambda_i$ is the lambda of the particle $p_i$ being in the cell at the time of traversal $t$, $\P{p_i\in c}$ depicts the probability that the particle $p_i$ is the one located in the cell and $\P{e_i \,\middle|\, p_i\in c}$ is the probability that the particle $p_i$ exists. 
	In the case where there is truly no obstacle in the cell, the probability $\P{p_i\in c}$ will be one for the static particle, which can have a very low lambda, thereby meaning that the cell is safe to cross, or a high lambda, meaning that the cell contains a static obstacle.
	Assuming that every cell has exactly one obstacle, we can compute the probability of the particle $p_i$ of being in the cell $c$ as the probability that only the particle $p_i$ creates a collision given that there is exactly one obstacle in the cell:
	\begin{equation}
		\begin{aligned}
				&\hskip-2em\P{p_i\in c \,\middle|\, (p_0\in c) \oplus \dots \oplus (p_{N_P-1}\in c)} = \\
						  & \frac{ (1 - \expp{-\lambda_i\area}) \prod_{j\neq i} \expp{-\lambda_j\area}}{\sum_k(1 - \expp{-\lambda_k\area}) \prod_{j\neq k} \expp{-\lambda_j\area}} \\
						&\propto \frac{1-\expp{-\lambda_i\area}}{\expp{-\lambda_i\area}},
		\end{aligned}
		\label{eq:update}
	\end{equation}
	for a cell $c$ of area $\area$ where $N_P$ particles lie in it, $\oplus$ being the standard XOR operator.
	The probability of existence $\P{e_i \,\middle|\, p_i\in c}$ is defined as the joint probability that the particle did not collide with the static environment since its creation and that it still follows an obstacle, computed as
	\begin{equation}
		\P{e_i \,\middle|\, p_i\in c} = \expp{-\sum_{c\in\mathcal{P}_i} \lambda_{c,s}} \cdot \expp{-\tau\cdot t_i},
	\end{equation}
	where $\mathcal{P}_i$ is the path (set of crossed cells) of the particle $p_i$, $\lambda_{c,s}$ is the lambda of the static particle at the cell $c$ (we allow dynamic particles to cross without collision), $t_i$ is the time since the last measurement `hit' (\ie the lidar hits the obstacle) of $p_i$ and $\tau$ is the rate of the distribution.
	Indeed, if the particle has not been measured for a long time, it is very likely that either the particle lost the obstacle or that the obstacle left the field of view of the robot.
	This probability can also take into account other sensors such as a camera: for instance, the probability of existence of particles of type `pedestrian' should drop if the camera informs the robot that there is a car at this position.

\subsection{Measuring particles}
Using the expressions of the probability of collision given by \autoref{eq:field} and the expectation of the intensities of the cells given by \autoref{eq:e_lambda}, we provide a means of estimating the lambdas of the static field as well as the dynamic particles using a lidar sensor.
		As in {\citep{Laconte2019}}, we determine the combination of the intensities $\lambda=\{\lambda_i\}$  of the particles that maximizes the expectation of the {$K$} beams the lidar has shot since the beginning.
		The lidar is modelled as a range sensor with an error region of area {$e$}: for a lidar measurement, the true position of the obstacle is contained within the error region, itself centered on the measurement.
In the case of a perfect sensor, this region is reduced to a point at the measurement position.
Using the derivation in {\cite{Laconte2019}} and the expectation of the intensity given by \autoref{eq:e_lambda}, 
	the intensities of the particles are given by
	\begin{equation}
			\lambda_i = \frac{1}{e}\ln\left(1+\frac{h_i}{m_i}\right),
			\label{eq:lambda}
	\end{equation}
	where {$h_i$} is the sum of probabilities $\P{p_i\in c_h}\P{e_i\,\middle|\,p_i\in c_h}$ each time the particle {$p_i$} has been counted as `hit' in the cell {$c_h$} (\ie was in the region of error of the sensor) and {$m_i$} the sum of probabilities $\P{p_i\in c_m}\P{e_i\,\middle|\,p_i\in c_m}$ each time the particle {$p_i$} has been counted as `miss' in the cell {$c_m$} (\ie the lidar beam crossed the cell without collision).
	For this equation to be true, one needs the assumption that for a given measurement, all of the particles measured in the error region of the sensor have the same intensity.
	As the lidar error region tends to be small, this assumption holds in every situation except for low-intensity static particles.
	Indeed, high-intensity dynamic particles can easily come to the cell containing the low-intensity particle, thereby the assumption that every obstacle has the same intensity no longer holds.
This case is tackled by assuming that the low-intensity static particle has the same intensity as the high-intensity particles, therefore overestimating its intensity and keeping a conservative approach.

\subsection{Particle Management}
	Using the particles update equation previously derived, the Lambda-Field is updated as described below.
	At each iteration, the particles evolve, are updated with lidar measurements and are resampled.
	First, the particles evolve using a simple update equation:
	\begin{equation}
	\begin{aligned}
		\bm{x}_i^t &= \bm{x}_i^{t-1} + \bm{v}_i^t\Delta t, \\
		\bm{v}_i^t &= \bm{v}_i^{t-1} + \bm{a}_i\Delta t \quad\text{with } \bm{a} \sim\mathcal{N}(0,\bm{\Sigma}_i),
	\end{aligned}
	\label{eq:evolution}
	\end{equation}
	for the particle $p_i$ of position and speed $\bm{x}_i^t, \bm{v}_i^t\in\R^2$ at the time $t$, where $\bm{a}_i\in\R^2$ is a centered Gaussian random variable of covariance $\bm{\Sigma_i}$ depicting the acceleration of the particle.
	Evidently, static particles have zero speed and acceleration. %
	Then, for each measured cell $c$, we compute the probability that the particle $p_i$ is indeed the particle in the cell.
	Each particle in the `hit' zone (resp. `miss' zone) of the lidar measurements increments its `hit' counter $h_i$ (resp. `miss' counter $m_i$) of the quantity $\P{p_i\in c}\P{e_i\,\middle|\, p_i\in c}$ as shown in \autoref{eq:lambda}.
	Moreover, particles have a low probability to switch classes (\ie `pedestrian' switching to `car' and vice versa).
	This consideration avoids an incorrect convergence of the particles (\eg a low velocity car can be represented as a group of pedestrians, however both hypotheses have to be maintained).

	Once the particles have been updated, they are resampled according to the joint probability of their existence and selecting one of the cells in which they are located: 
	\begin{equation}
		w_i \propto\, \P{p_i\!\in\! c}\P{ e_i \,\middle|\, p_i\in c} \!\sum_{p_i\in c} \!1\!-\!\expp{-\area\E{\lambda_c^t}}
	\end{equation}
	Moreover, particles can be born during the resampling step.
	Indeed, obstacles can appear on the map and the particles might not converge to an obstacle, leaving a dynamic obstacle unidentified.
	Even if this situation is less likely the more particles we have, dealing with such a critical case is necessary.
	Hence, we allow the birth of particles inside any cell $c$ measured `hit' by the lidar with
	\begin{equation}
		w_c^{\mathtt{birth}} = \gamma\cdot\expp{-\area\sum_{c\in\mathcal{E}_k}\E{\lambda_{c}^t}},
	\end{equation}
	for every `hit' measurement of error region $\mathcal{E}_k$, where $\gamma$ is a coefficient controlling the proportion of births at each resampling.
	If the measured cell is already populated by particles of high lambdas, meaning that the obstacle is already represented, the birth probability drops to zero.
	If the particle which is born is picked during the resampling, random class and speed are drawn from a uniform distribution.
	\autoref{alg:pm} summarizes the procedure.

	\vskip-.5em
\begin{algorithm}[htbp]
\SetAlgoLined
 \Repeat{True}{
	 Evolve particles using \autoref{eq:evolution} \\
	 Update particles with measurements using \autoref{eq:lambda} \\ 
	 \ForAll{particles $p_i$}{
		Draw a particle $p_k$ with weight $w_k$ and birth weight $w_c^{\mathtt{birth}}$ \\
		Replace particle $p_i$ with particle $p_k$ \\
	 }
  }
 \caption{Particles management}
 \label{alg:pm}
\end{algorithm}
\vskip-1em

	Once every particle has been updated and resampled, we infer the underlying distribution for each cell.	
	Indeed, planning while taking into account every particle would not reach real-time constraints.
	For each cell, the speed is modeled as a normal distribution $\mathcal{N}(\mu_v, \sigma_v)$ whereas the direction is modeled using the Von-Mises distribution (as the direction lies on a circle) of parameters $\mu_\theta, \kappa_\theta$, where these parameters are estimated as in \cite{Senanayake2018}.
	In the case where $\kappa_\theta$ is large enough, the Von-Mises distribution can be approximated by a normal distribution of the same mean with a standard deviation of $\sigma_\theta = \sqrt{1/\kappa_\theta}$.

	At the end of the mapping process, a grid is created where each cell contains an intensity for the static part, as well as an intensity for each type of dynamic particle (\ie summing the intensities of the particles of the same class, using \autoref{eq:e_lambda}).
	We do not sum the intensities of the different classes since the risk can depend on the type of obstacle.
	We instead provide each class intensity to the planner.
	As we defined three classes in this article (static, pedestrian and car), each cell contains three intensities, one for each obstacle.

\subsection{Risk assessment and Path Planning}
\begin{figure}[htbp]
	\centering
	\includegraphics[width=\linewidth]{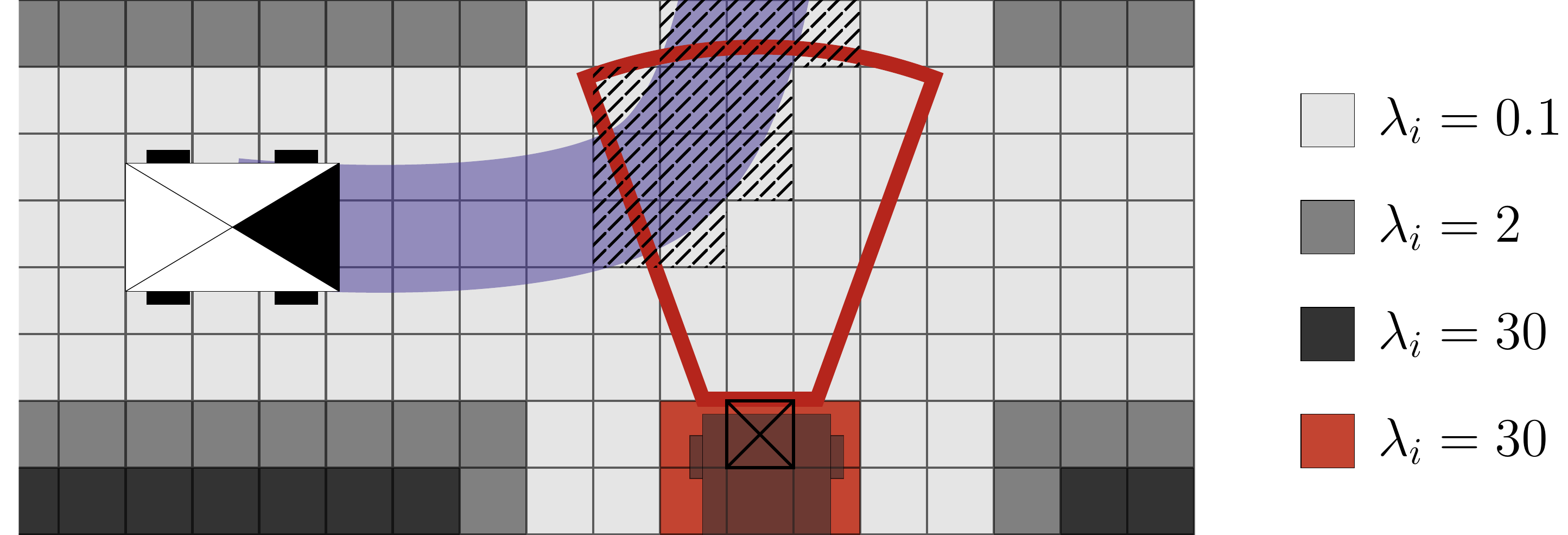}
	\caption{Example of risk assessment. The robot (left vehicle) wishes to cross the crossroad using the trajectory depicted in blue. 
			A dynamic obstacle arrives from the bottom of the map, with a probabilistic direction and speed, converted into a set (in red) that corresponds to the possible positions of the obstacle.
			Using these distributions for each dynamic cell (here depicted for the crossed one), the potential collision positions are marked (dashed cells).  
			If the times of traversal intersect, the obstacles are added to the crossed cells for risk assessment.
		}
	\label{fig:dlf_ra}
	\vskip-1em
\end{figure}

Using the data provided by the dynamic map, we generate safe trajectories for the robot.
As in \citet{Gerkey2008}, we sample commands of translational and rotational velocities $(v,w)$ and choose the one leading the closest possible to the objective while being below the allowed risk.
In each cell, the robot has to go through all of the obstacles simultaneously (\ie static obstacle and all of the dynamic ones arriving in the cell at the same time as the vehicle).
The probability of colliding with the obstacle $o_k$ in a cell $c$ containing $N_O$ obstacles $o_0, \dots, o_{N_O-1}$ of intensities $\lambda_0, \dots, \lambda_{N_O-1}$ at the time $t$ is the joint probability of not colliding the other obstacles $\{o_i\}_{i\neq k}$ and colliding with the obstacle $o_k$:

\begin{align}
	\P{\mathtt{coll}_{c,k}}	&=\int_0^{\area} \hskip-.5em\expp{-a\sum_{i\neq k}\lambda_i} \cdot\lambda_k\expp{-a\lambda_k}\dif a \notag\\
						&= \frac{\lambda_k}{\E{\lambda_c^t}} \left[1 - \expp{-\area\E{\lambda_c^t}}\right],
	\label{eq:coll_c}
\end{align}
where $\E{\lambda_c^t} = \sum_{i=0}^{N_O-1}\lambda_i$ is the expected lambda of the cell $c$ at the time of traversal $t$, where the intensities $\lambda_i$ of the incoming obstacles $o_i$ have been pondered by the probability of reaching the cell beforehand.
One can note that the expectation is the same as \autoref{eq:e_lambda} where we lifted the assumption that only one obstacle can be in the cell.
Indeed, even if only one obstacle can truly be in a cell, several obstacles can reach the cell from different directions and create a collision with the robot which attempts to cross it. %
Since several obstacles can hit the robot in the same cell, the assumption of each cell having only one obstacle is lifted, yielding $\P{p_i\in c}=1$ for the computation of expectation in risk assessment.
Using \autoref{eq:coll_c}, we compute the risk of a path as the expectation of the risk function $r(a, o_k)$, where $a$ is the traversed area at the time of the collision and $o_k$ is the obstacle the collision occurs with, as
\begin{align}
	\E{r(\cdot)} &= \sum_{i=0}^{N-1} K_i \sum_{o_k\in c_i} \frac{\lambda_k}{\E{\lambda_{c_i}^{t_i}}} \cdot r(i\area, o_k),\\
					\text{with }	K_i = &\expp{\!\!-\!\!\sum_{j=0}^{i-1}\area\E{\lambda_{c_j}^{t_j}}}\!\!\!\left(\!1\!-\!\expp{\!-\area\E{\lambda_{c_i}^{t_i}}}\!\!\right), \notag%
	\label{eq:risk}
\end{align}
for a path passing through the cells $\{c_0, \dots, c_{N-1}\}$ at the times $\{t_0, \dots, t_{N-1}\}$ of expected lambdas $\E{\lambda_{c_i}^{t_i}}$.
The curvilinear abscissa $s$ is linked to the traversed area by the relation $s = i\area / L$ where $L$ is the width of the robot.
In order to determine which obstacles are in the cell at the time of traversal, we convert the velocity distribution of each dynamic cell into a set as depicted in \autoref{fig:dlf_ra}. 
The set corresponds to the shape of the obstacle's (\ie dynamic cell) path using two sigmas on its velocity and orientation. %
This shape is then tested to cross the cells of the robot's path, using the Gilbert-Johnson-Keerthi distance algorithm.
In the case of a collision between the robot's path and the cell's path (dashed cells in \autoref{fig:dlf_ra}), the earliest arrival and latest departure time of the dynamic obstacle and the robot are compared. %
If the time intervals intersect, the obstacle is said to be in the cell of the robot's path. %
Using this risk assessment method, the robot is able to effectively assess the risk of a path.
In the following section, the risk is defined as the change of kinetic energy arising from the collision, thus taking into account the harmfulness of the collision for both the robot and the obstacle it collides with.

\section{Validation of the framework}
In order to show the applicability of our theory, we implemented our mapping framework on a Jetson TX2 GPU.
We used the robot depicted in \autoref{fig:intro} (right), equipped with a lidar LMS-151 located at its front.
As the aim is to plan short distance trajectories and the map is centered on the robot, we only used the odometry for the relative displacements of the robot.
To estimate the states of the dynamic obstacles, we used \num{2e4} particles in the experiments, running at more than \SI{10}{\hertz} and more than \SI{5}{\hertz} if the planning segment is carried out on the same GPU. %
The map size was set to \num{200}\,$\times$\,\num{200} cells of size \num{15}\,$\times$\,\num{15}\,\si{\cm}, resulting in a map of \num{30}\,$\times$\,\num{30}\,\si{\m}. 
In order to accelerate the convergence of the particles towards the obstacle, particle velocities were resampled with a Gaussian noise of $\sigma=\SI{.3}{\m\per\s}$.

\subsection{Evaluation}
\begin{figure}[thbp]
	\vskip-1em
	\centering
	\includegraphics[width=\linewidth]{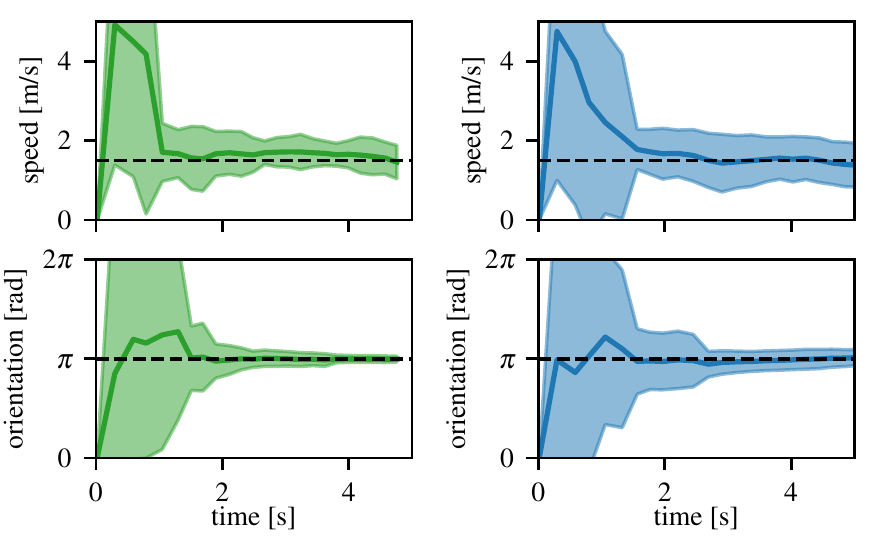}
	\vskip-1em
	\caption{Convergence of the speed and orientation of a single dynamic obstacle (Left: pedestrian; Right: car) where the ground truth is depicted in dashed black. The mean is displayed in solid line whereas the confidence interval at two sigmas is depicted in light shade.}
	\label{fig:convergence_velocities}
	\vskip-1em
\end{figure}

First, we show that the convergence speed of the framework.
Using the same type of validation as \citet{Nuss2016a}, we simulated an environment where an obstacle was approaching the robot at a known velocity and orientation.
This experiment was repeated 50 times in which the obstacle was either a pedestrian or a car.
The velocity of the obstacle was chosen to be {\SI{1.5}{\m\per\s}}, as this profile of speed can be matched by either the `pedestrian' or the `car' class.
Using the fact that only one dynamic obstacle is in the environment, we retrieve the mean and a confidence interval at two sigmas for the velocity and the orientation.
\mbox{\autoref{fig:convergence_velocities}} shows the results for a pedestrian (in green) and a car (in blue).
We can see that both the speed and the orientation converge to a valid value after less than {\SI{2}{\s}}.
Furthermore, as the framework assesses the risk by allowing two sigmas on the velocity and the orientation, the ground truth is always contained in the estimation.
At first, the speed of the obstacle is overestimated as both cars of high velocities and pedestrian of low velocities coexist.
In the case of the pedestrian, all particles of class `car' are discarded as their size is too large for the obstacle (\ie they lie in cells counted as free by the sensor).
Once the `car' particles are removed at \mbox{$t\approx\SI{1}{\s}$}, the mean velocity converges to the true value at the next iteration.
In the case of the car, the car particles of high velocities are discarded as soon as they leave the measurements zone, which is indicated by the fact that the convergence of the mean is smoother than in the pedestrian experiment.

\begin{figure}[htbp]
	\centering
	\includegraphics[width=\linewidth]{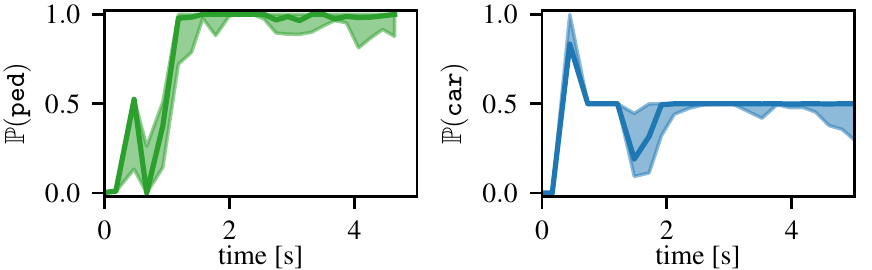}
	\vskip-1em
	\caption{Convergence of the class probability for a pedestrian (left) and a car (right) with the quantiles at 25\% and 75\% in light shade. Whereas the pedestrian has its class quickly inferred, the framework cannot decide between a car and a group of pedestrians in the case where the car has a speed matching both classes.}
	\label{fig:convergence_classes}
	\vskip-1em
\end{figure}

The convergence of the obstacle class is also studied.
We compute the probability of the obstacle to be of a certain class as the probability to collide with dynamic cells of the class.
\autoref{fig:convergence_classes} shows the resulting convergence for a pedestrian and a car.
In the case of the pedestrian, the framework quickly converges to the real class of the obstacle after \SI{1}{\s}.
However, in the case of the car, the framework cannot decide whether the obstacle is a car or a group of pedestrians. 
Indeed, with only a lidar, the obstacle can match both classes equivalently.
Other sensors such as a camera could remove the ambiguity.
Also, the class would be determined to be a car if the velocity profile only matched the `car' class (\ie the obstacle moves at greater velocity).
Therefore, the framework is able to effectively infer the velocity, orientation and class (when possible) of the different obstacles.

\subsection{Validation of the approach}

In the following experiment, the risk is defined as the maximum gain of kinetic energy between the robot and the obstacle it collided with, assuming inelastic collisions, as
\begin{align}
	&r(s, o_k) = \max\left(\frac12m_R \left(v_R(s)-v_f\right)^2, \, \frac12 m_k \left(v_k - v_f\right)^2\right)\notag \\
	&		  \text{with } v_f = \frac{m_Rv_R(s)+m_kv_k}{m_R+m_k}
\end{align}
where $m_R, v_R(s)$ is the mass and velocity of the robot at the curvilinear abscissa $s$, $m_k, v_k$ the mass and velocity of the obstacle $o_k$, and $v_f$ the final velocity of the obstacles after collision.
The mass of the robot was set to \SI{150}{\kg} while the masses of the `pedestrian' and `car' classes were set to \SI{80}{\kg} and \SI{500}{\kg} respectively.
This risk enables the consideration of both the possible damages suffered by the robot but also by the obstacle it collided with. 
As such, the decision to collide with a pedestrian takes into account that although the robot will suffer little damage, the pedestrian is at a much greater risk.
Note that the risk function can be adapted depending on the context and can take into account other elements if available such as slippage, car deformation and so on.
The static environment is assumed to have infinite mass, meaning that collisions with the static environment will always lead the vehicle to stop.
As shown in \autoref{fig:intro}, the robot had to move through an urban-like environment consisting of a crossroad, where other agents such as pedestrians and cars were also evolving.
A car was approaching in the other lane and a pedestrian entered the field of view of the robot from behind, afterwards crossing the road in front of it.
For obvious security reasons, in this experiment, the velocity of the robot was bounded such that it will always be able to instantly stop in the case of hazardous situations where every path is too risky, contrary to the scenario depicted in \autoref{fig:example_lf}.

\begin{figure}[t]
	\centering
	\includegraphics[width=\linewidth]{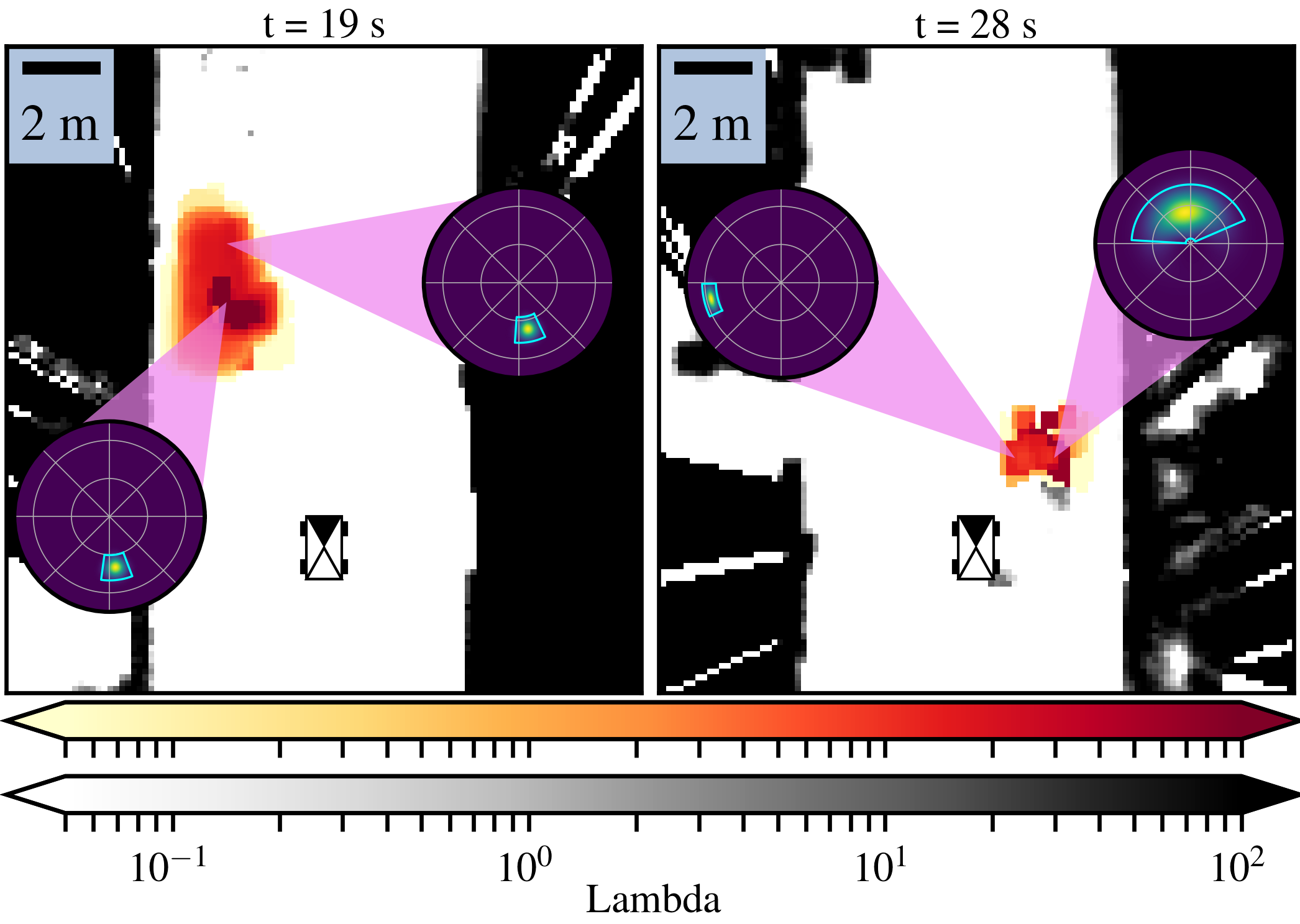}
	\vskip-1em
	\caption{Example of field resulting from our method, where the robot is located at the bottom center of the map.
			The static (resp. dynamic) environment is represented using a gray (resp. red) scale.
			A car is approaching the robot at $t=\SI{19}{\s}$, whereas a pedestrian is crossing the road at $t=\SI{28}{\s}$ right in front of the robot.
			Some speed distributions of the cells are displayed in polar plots (angle for the orientation, radius for the velocity), where inner circles correspond to a step of \SI{1}{\m\per\s}.
	}
	\label{fig:map_dlf}
	\vskip-1em
\end{figure}

\autoref{fig:map_dlf} shows the resulting Lambda-Field for two different timestamps.
The static environment is depicted with a gray scale, whereas the dynamic environment is shown with a red scale.
Using our framework, the velocity distribution of each cell is extracted as well as a lambda of the static environment and each type of particle (\ie in our case a lambda for the `car' class and a lambda for the `pedestrian' class).
At $t=\SI{19}{\s}$, the large dynamic high-lambda zone (in red) of the map corresponds to the car, where the probability of being a car is approximatively \SI{50}{\%} for the underlying cells.
The probability did not converge to \SI{100}{\%} because without additional sensors, the framework cannot decide whether the obstacle is a car or a group of pedestrians, as the velocity of \SI{1.2}{\m\per\s} is possible for both classes.
At $t=\SI{28}{\s}$, a pedestrian emerges from behind the robot, then crosses the road right in front of it a few seconds later.
When the robot detects the new obstacle, an inconsistency between the map and the measurements is found, leading the framework to create many particles on this location.
Note the light gray trace which is left behind the pedestrian, as the framework had not yet decided whether the obstacle was static or dynamic.
As shown in the right polar distribution (\ie angle for the orientation, radius for the velocity), the hypothesis of the pedestrian pursuing its northward trajectory (top of the figure) is maintained.
This hypothesis models the fact that without other sensors, the obstacle can in fact represent two pedestrians walking together, with the probability that they can change directions.
The wrong hypothesis (\ie pedestrian moving north) is discarded over the next iterations.

\begin{figure}[t]
	\centering
	\includegraphics[width=\linewidth]{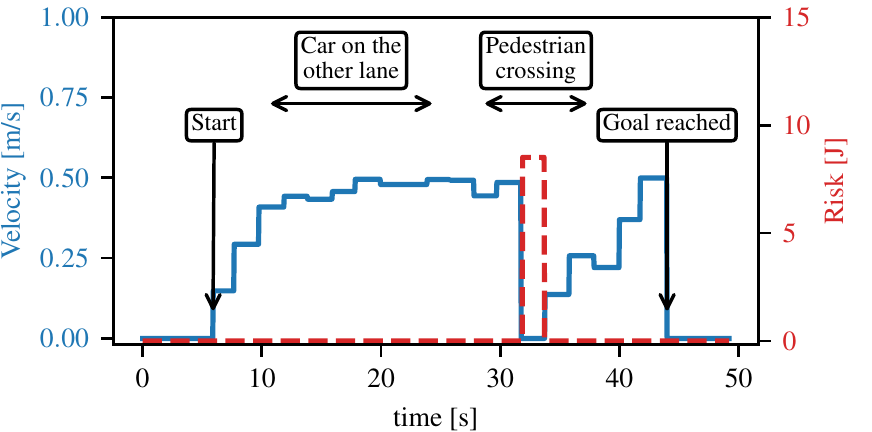}%
	\vskip-1em
	\caption{Risk undergone by the robot during its traversal with the associated speeds.
		First, a car passed the robot in the other lane, where its speed was precise enough not to cause the robot to brake or change direction.
		After that, a pedestrian emerges from behind the robot and crosses the road in front of it, leading the robot to stop and wait for the pedestrian to free the way.
		The robot then rejoins its goal without further obstacles.
	}
	\label{fig:risks}
	\vskip-2em
\end{figure}

Finally, we used this dynamic map to plan safe paths for the robot.
The goal of the robot was set at the top of the map, \SI{15}{\m} away from its position. %
To do so, we set the maximum risk to be $r_{\text{max}}=\SI{1}{\J}$ where any path below this risk is considered safe. %
\autoref{fig:risks} shows the risks the robot underwent during the traversal.
First, no obstacles were in sight, leading the robot to accelerate to its maximum speed, here at $\SI{0.5}{\m\per\s}$.
At $t=\SI{10}{\s}$, a car entered the field of view of the robot.
As the obstacle was far away enough during the convergence of the speed and orientation, the robot did not stop its course.
At $t=\SI{22}{\s}$, the car passed on the left side of the robot.
As the velocities were precise enough not to encounter the path of the robot, it continued its course at full speed.
At $t=\SI{31}{\s}$, the pedestrian took a hard left turn, deviating from its expected trajectories thereby leading the framework to birth particles on its position.
Consequently, the robot detected a danger on every path it could take, leading it to stop as this decision is the one minimizing the risk.
The associated risk is then the risk of the pedestrian running into the robot, thus harming himself.
In contrast, the Bayesian occupancy grid would only yield a probability of collision (in this case equal to one) and the robot could not distinguish between the collision at full speed and the collision at rest since both paths lead to a collision with the pedestrian.
If the velocity of the robot did not allow it to stop, the robot would then prefer to collide with a parked car of same weight, as the resulting risk is lower.
After a few iterations, the velocity of the pedestrian re-converged, and the robot continued on its way as soon as the pedestrian left.
The robot accelerated to its maximum velocity and reached its goal safely.

\vspace{-.3em}
\section{Conclusion}
\vspace{-.2em}
In this article, we proposed a novel framework for generic risk assessment in occupancy grids.
Using particles, we modeled both static and dynamic environments, deriving at the same time the nature of the obstacles.
We first showed how to compute the Dynamic Lambda-Field using lidar measurements.
Then, the resulting dynamic map is used to assess the risk for a given path, here defined as the change of kinetic energy due to a collision with an obstacle.
On the contrary, the Bayesian occupancy grid would only yield the probability of collision, hence not being able to differ between a collision with a car or a pedestrian.
Using this formulation of the risk, the robot was able to plan real-time safe trajectories in a dynamic environment.
As this work focused on explaining the theoretical framework and providing a use case for its application, future work will involve extensive experiments with the framework in both real and simulated benchmarks, allowing for more complex and hazardous scenarios.
Future work will also address particle convergence guarantees for a wider range of scenarios.
Finally, other risks, metrics and particle types will be used to better characterize the risk of a path in more complex situations.

\vspace{-.5em}
\section*{ACKNOWLEDGMENT}
\vspace{-.3em}
We thank Ted Morell for his help with the implementation of the framework.
This work was sponsored by a public grant overseen by the French National Research Agency as part of the “Investissements d’Avenir” through the IMobS3 Laboratory of Excellence (ANR-10-LABX-0016), the IDEX-ISITE initiative CAP 20-25 (ANR-16-IDEX-0001) and the RobotEx Equipment of Excellence (ANR-10-EQPX-0044).
\renewcommand*{\bibfont}{\small}
\vskip-1em
\printbibliography

\end{document}